# A Specialized Large Language Model for Clinical Reasoning and Diagnosis in Rare Diseases

Tao Yang[1,#], Dandan Huang[2,#], Yunting Lin[1,#], Pengfei Wu[1], Zhikun Wu[1], Gangyuan Ma[3], Yulan Lu[1], Xinran Dong[4], Dingpeng Li[2], Junshuang Ge[1], Zhiyan Zhang[1], Xuanzhao Huang[1], Wenyan Nong[1], Yao Zhou[1], Hui Tang[1], Hongxi Yang[2], Shijie Zhang[2], Juan Li[1], Xiaojun Cao[1], Lin Yang[4], Xia Gao[1], Kaishou Xu[1], Xiaoqiong Gu[1], Wen Zhang[1], Huimin Xia[1], Li Liu[1,*], Wenhao Zhou[1,*], Mulin Jun Li[1,*]

[1]Guangzhou Women and Children's Medical Center, Guangzhou Medical University, Guangzhou, China.
[2]Center for Cardiovascular Diseases, The Province and Ministry Co-sponsored Collaborative Innovation Center for Medical Epigenetics, School of Basic Medical Sciences, Tianjin Medical University, Tianjin, China.
[3]Guangzhou National Laboratory, Guangzhou, China.
[4]Center for Molecular Medicine, Children's Hospital of Fudan University, Shanghai, China.

[#]The authors contributed equally to this work
*Correspondence: mulinli@connect.hku.hk (M.J.L), zhouwenhao@fudan.edu.cn (W.H.Z), liliuchina@qq.com (L.L)

**Abstract:**

Rare diseases affect hundreds of millions worldwide, yet diagnosis often spans years. Convectional pipelines decouple noisy evidence extraction from downstream inferential diagnosis, and general/medical large language models (LLMs) face scarce real-world electronic health records (EHRs), stale domain knowledge, and hallucinations. We assemble a large, domain-specialized clinical corpus and a clinician-validated reasoning set, and develop RareSeek-R1 via staged instruction tuning, chain-of-thought learning, and graph-grounded retrieval. Across multicenter EHR narratives and public benchmarks, RareSeek-R1 attains state-of-the-art accuracy, robust generalization, and stability under noisy or overlapping phenotypes. Augmented retrieval yields the largest gains when narratives pair with prioritized variants by resolving ambiguity and aligning candidates to mechanisms. Human studies show performance on par with experienced physicians and consistent gains in assistive use. Notably, transparent reasoning highlights decisive non-phenotypic evidence (median 23.1%, such as imaging, interventions, functional tests) underpinning many correct diagnoses. This work advances a narrative-first, knowledge-integrated reasoning paradigm that shortens the diagnostic odyssey and enables auditable, clinically translatable decision support.

**Introduction**

Rare diseases, typically defined by a prevalence below 1/2000[1,2], now encompass well over 10,000 distinct disorders and affect hundreds of millions worldwide, yet patients commonly endure a prolonged "diagnostic odyssey," with the average time for an accurate diagnosis is 4-8 years[3]. Contemporary differential diagnostics seek to integrate longitudinal clinical evidence, for example standardized phenotypes extracted from highly heterogeneous electronic health records (EHRs)[4] and clinically pathogenic variants/genes prioritized by high-throughput family-based sequencing[5,6], thereby incrementally improving diagnostic yield. In practice, achieving efficient and accurate clinical diagnosis still hinges on two tightly coupled steps: (1) faithful acquisition and standardization of complete clinical evidence, especially phenotype terms formulated as Human Phenotype Ontology (HPO)[7] and molecular diagnostic reports (such as pathogenic/likely pathogenic variants documented in ClinVar[8]), and (2) rigorous diagnostic reasoning grounded in the patient's causal evidence chain, culminating in clinical decision-making supported by computational algorithms[9]. However, nearly all state-of-the-art clinical workflows and computational decision-support strategies operationalize these two challenging tasks in isolation, treating key phenotypes/variants extraction as a pre-processing task and downstream inference as a largely separate ranking exercise[10]. This structural decoupling injects biases and uncertainty at each step, leaving the diagnostic odyssey largely unabated.

Large language models (LLMs) are advancing rapidly in medicine[11,12]. General models, including GPT series[13], LLaMA[14], PaLM[15], and DeepSeek[16], increasingly support clinical tasks. Building on these, specialized medical LLMs have emerged, such as AMIE[17], MedPaLM-2[18], MedFound[19], Baichuan-M2[20], MMedLM[21], Meditron[22] and Clinical Camel[23]. Collectively, these systems demonstrate strong performance in medical QA and decision support, highlighting the promise of domain adapted LLMs for differential diagnostics of rare disease. Pioneer attempts to apply LLMs to rare disease diagnosis show mixed, often contradictory results. Some studies show GPT-4/4o outperforming general baselines and, in limited settings, even surpassing existing phenotype-driven pipelines[24]. Other studies report that general LLMs, such as OpenAI o1 and GPT-4o, underperform bioinformatic tools like Exomiser when limited to extracted HPO terms[25]. Agentic hybrids like DeepRare yield modest gains yet remain dependent on accurate phenotype extraction and normalization[26]. These inconsistencies leave an open question on whether LLMs can truly surpass current strategies for rare disease differential diagnosis.

Training domain-specialized LLMs for rare diseases faces intertwined data and technical challenges. First, data scarcity and distributional mismatch limit both training and evaluation. Available clinical corpora are relatively small, fragmented, and heterogeneously labeled, while public benchmarks skew toward structured phenotypes rather than full clinical narratives, which inflates performance on curated inputs and undermines external validity in real clinics[24]. Second, phenotype-centric pipelines compress clinical evidence and lose diagnostic signal. Phenotype extraction and HPO normalization fail to capture non-phenotypic cues relevant to rare disease inference, such as pathology and histology results, environmental and lifestyle factors, exposure history, and clinical interventions[27]. Moreover, well-annotated HPO-mapped resources such as Phenopacket[28] and RareBench[24] are predominantly retrospective, ground-truth datasets that do not reflect the difficulty, incompleteness, and noise of evidence acquisition during real-time diagnosis. Third, LLMs may hallucinate findings and struggle with causal and temporal reasoning in the absence of latest variant-gene-phenotype-disease relationships[24,29], especially when evidence is incomplete. Together these challenges and remedies chart a course beyond current best clinical practices toward reliable and scalable rare disease diagnosis with LLMs.

In this study, we systematically assembled three rare disease-focused resources, including a large-scale clinical corpus (RareMed-Corpus), a clinician-validated diagnostic reasoning dataset (RareMed-CoT), and a graph-grounded retrieval resource (RareMed-RAG). Building on this foundation, we developed RareSeek-R1, a domain-specialized LLM tailored for rare disease diagnostic reasoning. The model employs a Progressive Parameter-Efficient Transfer Learning strategy in three stages[30]. First, domain-specific instruction tuning on the DeepSeek-R1 injects rare disease knowledge while preserving general linguistic competence. Second, chain-of-thought (CoT) fine-tuning strengthens multi-step clinical reasoning and integration of heterogeneous evidence. Third, integrative reasoning couples retrieval-augmented generation (RAG) with a curated rare disease knowledge graph (KG), aligning variant-gene-phenotype-gene-disease relations to improve factual fidelity, calibration, and interpretability. We then benchmarked RareSeek-R1 against leading LLMs and state-of-the-art phenotype-driven tools across five public and multi-center datasets, forming the largest evaluation to date in rare disease diagnosis. RareSeek-R1 consistently delivered superior accuracy and remained robust under noisy or atypical phenotypes. Human-AI comparative studies and AI-assisted trials in real clinical workflows confirmed its applicability and reliability for decision support. Finally, through an LLM clinical readiness assessment and interpretability analyses, we characterized RareSeek-R1's reasoning process and rationality, elucidated its internal inference mechanisms, and identified potential missing elements in conventional rare disease differential diagnostics pipelines.

## Results

### Large-scale multidimensional rare disease knowledgebase integration and RareSeek-R1 model development

RareSeek-R1 was trained through a systematic, three-stage pipeline that integrates large-scale data collection, standardized curation, and graph-grounded digitization (Fig. 1a). In stage one, domain-specific instruction tuning on RareMed-Corpus (149,341 independent rare disease-oriented texts, ~500M tokens) injected consolidated rare disease knowledge while preserving general language capability endowed by the DeepSeek-R1-70B base model. The rare disease-focused corpus comprised 48,852 de-identified, definitively diagnosed EHRs, 35,722 medical texts and guidelines, 30,101 PubMed case reports, and 34,666 synthetic cases, ensuring diversity and clinical relevance. In stage two, CoT fine-tuning on RareMed-CoT, which contains 17,477 high-quality reasoning chains, strengthened multi-step reasoning and evidence integration. This training shifted the model from outcome-only prediction to process-explainable diagnosis and improved robustness and interpretability. In stage three, GraphRAG (Graph Retrieval-Augmented Generation) -based integrative reasoning reduced hallucinations and aligned inference with up-to-date biomedical knowledge by coupling retrieval-augmented generation to a curated rare disease KG. RareMed-RAG harmonizes definitions and relations from commonly-used resources, such as ClinVar[8], HGMD[31], HPO[7], OMIM[32] and Orphanet[33], enabling precise traversal of variant-gene-phenotype-disease links (see details in Methods).

To enable a systematic evaluation of RareSeek-R1, we assembled a unified suite of five high-quality, digitized benchmarks, the largest rare disease diagnostic cohort to date (Fig. 1b). These include: 1) EHR-Internal includes 4,306 de-identified, definitively diagnosed clinical narratives from the same institutions as the training data, supporting internal validation on matched settings. 2) EHR-External includes 283 multi-center cases from independent hospitals, enabling out-of-distribution testing across institutions. 3) RareBench includes 1,197 public phenotype-only cases drawn from sources such as MME[34], LIRICAL[35], HMS[36], RAMEDIS[37], and PUMCH_ADM[24]; cases are standardized to HPO terms with OMIM/Orphanet diagnoses, assessing generalization on curated phenotypic profiles. 4) MedEHR-Variant includes 147 cases pairing full EHRs with whole-exome sequencing (WES), enabling joint clinical-genetic inference and quantifying the incremental value of genomic data and structured knowledge augmentation. 5) Phenopacket-Store includes 5,213 GA4GH-conformant, case-level records following the Phenopacket schema, with harmonized HPO phenotypes, variant annotations, and reference diagnoses, further supporting evaluation of phenotype- and genotype-aware pipelines (see details in Methods).

By leveraging above large and high-quality benchmarks, we evaluated RareSeek-R1 against five general LLMs (LLaMA 3.3-70B[38], GPT-4o[13], GPT-5[39], OpenAI o1[40,41] and QwQ-32B[42]), six medical LLMs (including Meditron-70B[22], MMedLM-70B[21], Clinical Camel-70B[23], HuatuoGPT-o1-70B[43], Baichuan-M2[20] and Baichuan-M1[44]), and five phenotype-driven tools (including Exomiser[5], PhenoDP[45], Phen2Disease[46], Base_IC[46] and Phrank[47]) across differential diagnostic accuracy, sensitivity to phenotypic noise/complexity, sequential clinical evidence value, gains from GraphRAG and genomic data, human-AI comparative performance, and interpretability/clinical readiness (Fig. 1c).

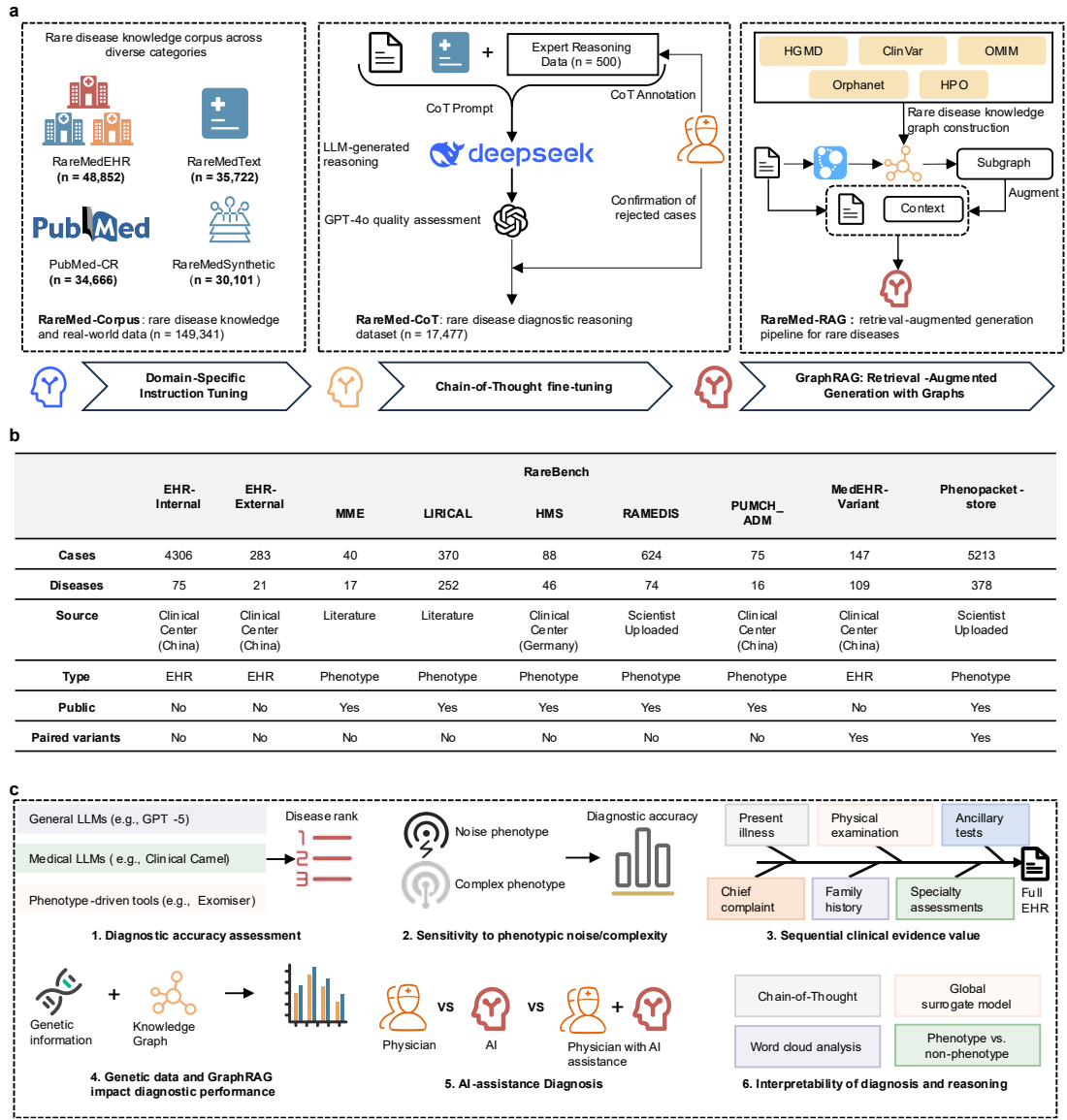

**Fig. 1: Schematic illustration of the development and evaluation of the RareSeek-R1 framework for rare disease diagnosis. (a)** Model development. RareSeek-R1 was built upon DeepSeek-R1-Distill-LLaMA-70B through domain-specific instruction tuning on RareMed-Corpus, followed by chain-of-thought fine-tuning on the RareMed-CoT dataset to improve diagnostic reasoning in rare diseases. The model further incorporated GraphRAG-based integrative reasoning grounded in a curated rare disease knowledge graph to ensure factual consistency and biomedical alignment. **(b)** Benchmark datasets. Comprehensive multi-source benchmark datasets were established to evaluate diagnostic performance across diverse case distributions, data origins, disease categories, and genetic annotation statuses. **(c)** Clinical evaluation of the AI system. The model was systematically evaluated for diagnostic accuracy,

robustness to phenotypic variability, reasoning with sequential clinical evidence, benefits from GraphRAG and genomic information, human-AI performance, and clinical interpretability.

**Comprehensive evaluation of rare disease differential diagnosis across LLMs and conventional methods**

To ensure fair and reproducible assessment, we implemented a unified evaluation protocol. For each case, models generated up to 20 candidate diagnoses, subsequently normalized to Orphanet identifiers via the MONDO API[48]. Accuracy was computed at multiple rank cutoffs and at two levels: (i) exact, when the predicted Orphanet code matched the reference; and (ii) hierarchical, when the prediction mapped to a parent concept of the reference diagnosis. Given the clinical actionability of higher-level categories in rare disease workflows, hierarchical matches were treated as correct in secondary analyses. This protocol was applied uniformly across all models and baselines (see details in Methods). Across EHR-Internal, EHR-External, and RareBench benchmarks emphasizing full EHR narratives or core phenotypes, DeepSeek-R1 achieved the strongest overall diagnostic performance among foundation models. This superior performance motivated our selection of DeepSeek-R1 as the base model for developing RareSeek-R1.

We first evaluated RareSeek-R1 against multiple baselines and competing tools on the EHR-Internal test set. RareSeek-R1 demonstrated the best overall diagnostic performance across all methods, with a Top-1 accuracy of $0.684 \pm 0.014$, significantly outperforming both general LLMs and phenotype-based diagnostic tools (Fig. 2a). Generally, Reasoning-enhanced LLMs outperformed general-purpose models, OpenAI o1 achieved the highest accuracy among non-specialized models ($0.487 \pm 0.016$), followed by Baichuan-M2 ($0.438 \pm 0.017$). Comparable trends were observed under hierarchical evaluation. Across 28 Orphanet-defined disease categories (noting that a single disease may map to multiple categories), RareSeek-R1 demonstrated consistently balanced performance, achieving > 0.80 accuracy in rare neoplastic or tumoral, rheumatologic, and urogenital diseases, but lower accuracy in rare immune (0.286), metabolic (0.498), and infertility disorders (0.477), delineating its current diagnostic boundaries.

Because conventional rare disease diagnostic tools largely rely on phenotype extraction from clinical evidence, particularly EHRs, we applied three established extractors

(PhenoTagger[49], PhenoBERT[4], ClinPhen[50]) for HPO normalization and compared their downstream differential diagnostic performance. Using PhenoTagger for HPO extraction yielded the highest accuracy across five downstream phenotype-based diagnostic methods (Exomiser, PhenoDP, Phen2Disease, Base_IC, Phrank) under both exact and hierarchical criteria. Accordingly, we used PhenoTagger to extract HPO terms for all subsequent analyses. Among conventional phenotype-driven approaches, Exomiser showed the highest accuracy (Top-1 = 0.057 ± 0.008; Top-3 = 0.138 ± 0.006), while PhenoDP and Phen2Disease performed similarly or worse, indicating that these approaches struggle without expert-curated HPOs and lack robust direct EHR parsing. To quantify the value of clinical context, we re-evaluated RareSeek-R1 on EHR-Internal using phenotype-only input. Compared with a Top-1 accuracy of 0.684 ± 0.014 using full text, phenotype-only input achieved a markedly lower Top-1 accuracy of 0.192 ± 0.012 (Fig. 2b). This gap suggests routine EHRs contain clinically decisive signals that automated HPO pipelines often miss or cannot map (e.g., personal genetic findings, disease trajectory, treatment response, longitudinal investigations, family history, imaging and laboratory evidence, and therapeutic exposures).

To evaluate robustness across heterogeneous data sources, RareSeek-R1 was further tested on the EHR-External dataset comprising multi-center real-world cases. RareSeek-R1 consistently outperformed all baselines, achieving a Top-1 accuracy of 0.719 ± 0.025, demonstrating strong generalization beyond the training distribution (Fig. 2c). In comparison, GPT-5 achieved 0.578 ± 0.004, while phenotype-based tools such as Exomiser showed limited performance (Top-1 accuracy of 0.046 ± 0.035). Across disease categories, RareSeek-R1 reached > 0.80 accuracy in rare bone, developmental defect, genetic, and ophthalmic disorder. On the independent RareBench[24] dataset curated around core phenotypes rather than full narratives, RareSeek-R1 achieved 0.392 ± 0.025 Top-1 accuracy (Fig. 2d), surpassing GPT-5 (0.353 ± 0.049). Among phenotype-based methods, Exomiser performed best (Top-1 = 0.232 ± 0.016), surpassing several medical LLMs (e.g., Meditron-70B, MMedLM-70B, Clinical Camel-70B). Notably, its Top-10 accuracy (0.575 ± 0.022) was comparable to RareSeek-R1 (0.577 ± 0.027). Across disease categories, RareSeek-R1 achieved >0.65 accuracy in rare gynecologic or obstetric, infertility, cardiac surgical, and urogenital diseases. Because most RareBench cases contain only key phenotypic information, this benchmark, unlike

EHR-Internal and EHR-External which emphasize comprehensive clinical narratives, further illustrates RareSeek-R1's stable performance across heterogeneous data regimes.

Our training framework adopts a Progressive Parameter-Efficient Transfer Learning paradigm, enabling stepwise knowledge infusion and targeted adaptation for rare disease diagnosis. To determine whether domain-specific instruction tuning effectively injects rare disease knowledge and whether CoT-based integration of heterogeneous evidence further enhances LLM diagnostic reasoning, we conducted component-wise ablations by systematically fine-tuning publicly available general and medical LLMs. As shown in Fig. 2e, all models demonstrated consistent performance gains across datasets after progressive fine-tuning. Following the rare disease knowledge infusion stage, the models achieved average improvements in diagnostic accuracy of 11.8%, 9.6%, and 4.4% on EHR-Internal, EHR-External, and RareBench, respectively. Building upon these gains, CoT fine-tuning, designed to enhance reasoning coherence and diagnostic reliability, yielded an additional average gain of 9.5%, 7.9%, and 3.9% across the three datasets. In summary, these results demonstrate that our stepwise training framework markedly strengthens LLM reasoning robustness and cross-domain generalizability in rare disease diagnosis.

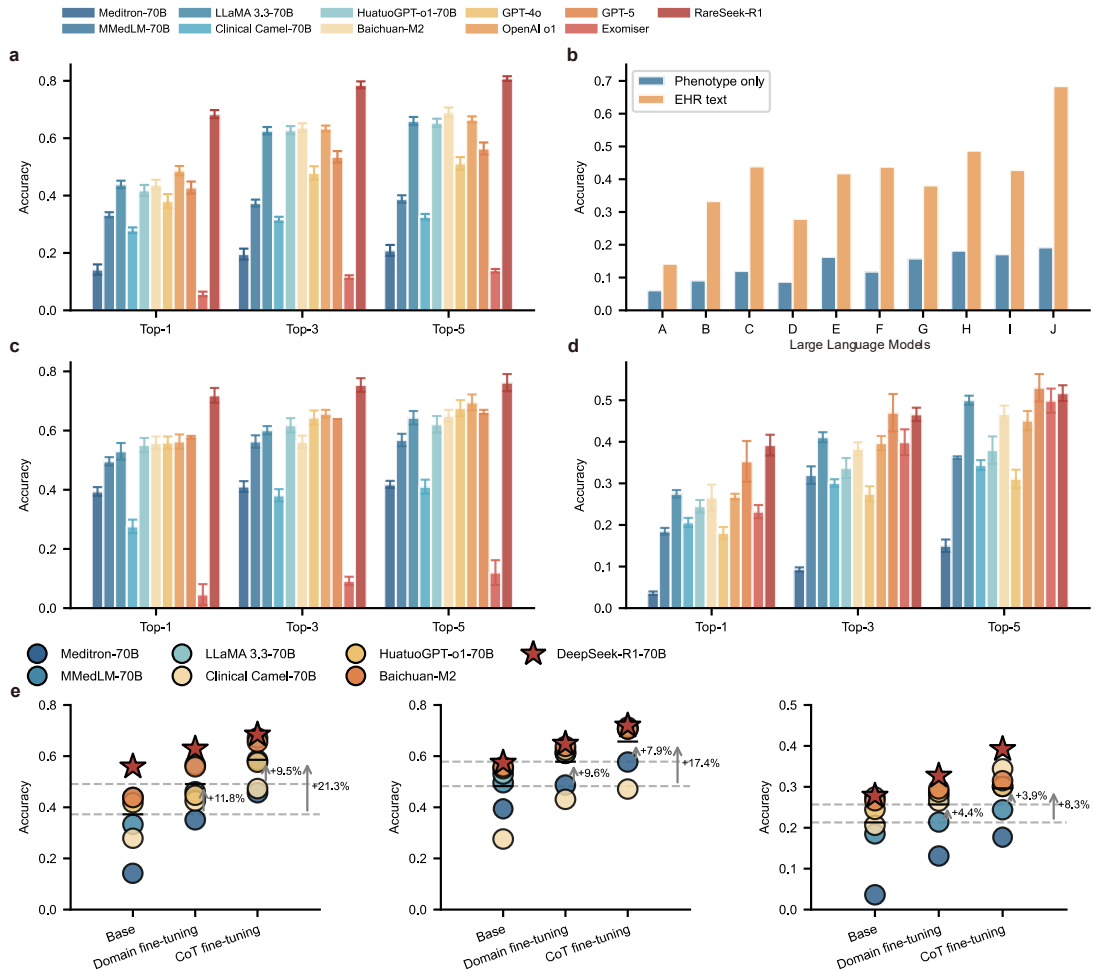

**Fig. 2: Performance analysis of rare disease diagnostic methods and LLM training strategy components across datasets. (a)** Comparative diagnostic performance of rare disease methods on EHR-Internal. **(b)** Diagnostic performance of LLMs comparing diagnoses based on phenotype only and full EHR narratives (A: Meditron-70B; B: MMedLM-70B; C: LLaMA 3.3-70B; D: Clinical Camel-70B; E: HuatuoGPT-o1-70B; F: Baichuan-M2; G: GPT-4o; H: OpenAI o1; I: GPT-5; J: RareSeek-R1). **(c, d)** Comparative diagnostic performance of rare disease methods on EHR-External **(c)** and RareBench **(d)**. **(e)** Contribution of LLM training components to diagnostic performance across datasets (left: EHR-Internal; middle: EHR-External; right: RareBench). The short horizontal line represents the mean performance across models. Percentage increases indicate improvements achieved through knowledge infusion fine-tuning and CoT fine-tuning.

**Impact of phenotypic noise and complexity on model performance**

In the aforementioned analysis, we found that when all EHR-extracted phenotypes were used, RareSeek-R1 and most LLMs sustained strong differential diagnostic performance on both routine EHR narratives and phenotype-dominant cases, whereas traditional phenotype-driven tools such as Exomiser were competitive with some general and medical LLMs only when phenotypic signals were highly specific and well curated, as in RareBench (Fig. 2d). This observation prompted two questions: whether conventional phenotype extraction and diagnostic pipelines lack the capacity to automatically process complex and heterogeneous clinical narratives due to limits in phenotype granularity, ontology coverage, and contextualization, and whether full clinical narratives contain additional unmapped phenotypes or non-HPO signals that materially facilitate correct diagnosis. To investigate these issues, we quantified within clinical EHRs the proportion of key phenotypes, defined as features that exactly match HPO terms for the confirmed diagnosis, and evaluated their impact on differential diagnostic accuracy. Given its size and disease diversity, all subsequent phenotype-based analyses were conducted on EHR-Internal (n = 4,306).

Using Human Phenotype Ontology Annotation (HPO-A), we defined key phenotypes as EHR-extracted features that exactly matched HPO-A annotations for the confirmed diagnosis. Cases were stratified by the per-patient count of key phenotypes into three tiers: Few (0-1, N = 2,368), Moderate (2-4, N = 1,675), and Rich (>4, N = 263). The Top-5 diagnostic accuracy of LLMs (e.g., RareSeek-R1, Baichuan-M2, OpenAI o1) increased with the number of key phenotypes, whereas traditional phenotype-driven methods (e.g., Exomiser) declined slightly, attaining only 0.145, 0.134, and 0.134 across tiers, versus RareSeek-R1 at 0.787, 0.819, and 0.840 (Fig. 3a). Notably, the proportion of key phenotypes among all extracted phenotypes decreased as tier increased (Few: $0.312 \pm 0.011$; Moderate: $0.235 \pm 0.011$; Rich: $0.136 \pm 0.009$; Fig. 3b), indicating that noise or unmapped phenotypes are pervasive in real-world EHRs and may constitute a larger share in records that also contain more key phenotypes.

To assess RareSeek-R1's differential diagnostic capability under well-annotated, HPO-mapped conditions, we re-evaluated performance on EHR-Internal using only key phenotypes. Under this setting, Exomiser's Top-5 accuracy improved markedly to 0.127, 0.369, and 0.427 across the Few, Moderate, and Rich tiers, respectively (Fig. 3c). This improvement indicates that Exomiser's diagnostic performance is highly sensitive to phenotype quality, with noisy or imprecise HPO terms substantially diminishing its accuracy. By comparison, RareSeek-R1 achieved Top-5 accuracies of 0.401, 0.721, and 0.734 using only key phenotypes, still largely outperforming phenotype-driven methods, yet remaining below its performance with full EHR narratives (Fig. 3d). These findings indicate that RareSeek-R1 perform well when restricted to key HPO-mapped features, but they derive additional gains from the richer clinical context present in unstructured EHR text.

Real-world EHRs for rare disease patients frequently contain low-information phenotypes that recur across many conditions and have limited discriminative value. We defined cases with highly overlapping or low-information features as complex-phenotype cases and constructed two cohorts, including the 300 cases with the lowest mean information content (Mean IC) and the 300 with the highest composite difficulty scores (see details in Methods). RareSeek-R1 achieved the highest accuracy in both cohorts (Top-1 = 0.723 and 0.520; Fig. 3e). Under the Mean IC criterion, Baichuan-M2 ranked second (Top-1 = 0.510); under the composite difficulty criterion, OpenAI o1 ranked second (Top-1 = 0.423). These findings delineate graded diagnostic complexity and underscore the robustness and generalizability of LLMs to overlapping, ambiguous phenotypes in real-world settings. Taken together, conventional phenotype extraction/normalization and phenotype-driven tools falter on complex, heterogeneous narratives. By reasoning directly over raw EHR text and leveraging stepwise rare disease knowledge infusion with CoT-based integrative reasoning, RareSeek-R1 achieves state-of-the-art performance across diverse benchmarks.

**Sequential effect and importance of clinical evidence categories for rare disease diagnosis**

In rare disease diagnosis, clinical evidence is accumulated sequentially rather than obtained in a single encounter, typically beginning with the chief complaint, followed by the history of present illness and family history, then proceeds through multiple rounds of physical examination, specialty assessments, molecular testing, and ancillary investigations. As information accrues, diagnostic judgments are progressively refined. To quantify the contribution of each EHR category to model performance of RareSeek-R1, we conducted an Incremental Information Addition Analysis and single-field ablations. In the incremental analysis, using the chief complaint as baseline, Top-1 accuracy was $0.475 \pm 0.012$; sequentially adding history of present illness, family history, physical examination, specialty assessments, and ancillary tests yielded Top-1 accuracies of 0.561 ± 0.009, 0.520 ± 0.015, 0.543 ± 0.020, 0.523 ± 0.009, and 0.684 ± 0.014, respectively (left panel of Fig. 3f). In the ablation experiments, starting from the complete EHR (Top-1 = 0.684 ± 0.014), removing individual fields produced the largest drops when omitting the chief complaint or ancillary examinations (0.463 ± 0.015 and 0.523 ± 0.017), followed by history of present illness, family history, physical examination, and specialist examination (0.552 ± 0.015, 0.567 ± 0.017, 0.579 ± 0.014, and 0.578 ± 0.017; right panel of Fig. 3f). These results indicate that the chief complaint, present illness, and ancillary tests are the most critical clinical categories for rare disease diagnosis, whereas family history, physical examination, and specialty assessments provide complementary value.

Overall, the superior performance of complete EHR-based diagnosis reflects not only the sequential accrual of clinical evidence but also the synergistic contribution of heterogeneous signals that RareSeek-R1 can exploit. Crucially, richer early-visit information enables earlier, more confident triage and differential diagnosis, helping to shorten the prolonged diagnostic odyssey in rare diseases. Moreover, key attributes, such as temporal course, demographics, treatment response, longitudinal workups, family history, exposures, and detailed tests, are often absent from or poorly captured by standardized HPO representations, limiting phenotype-driven tools compared with models reasoning directly over full EHR narratives.

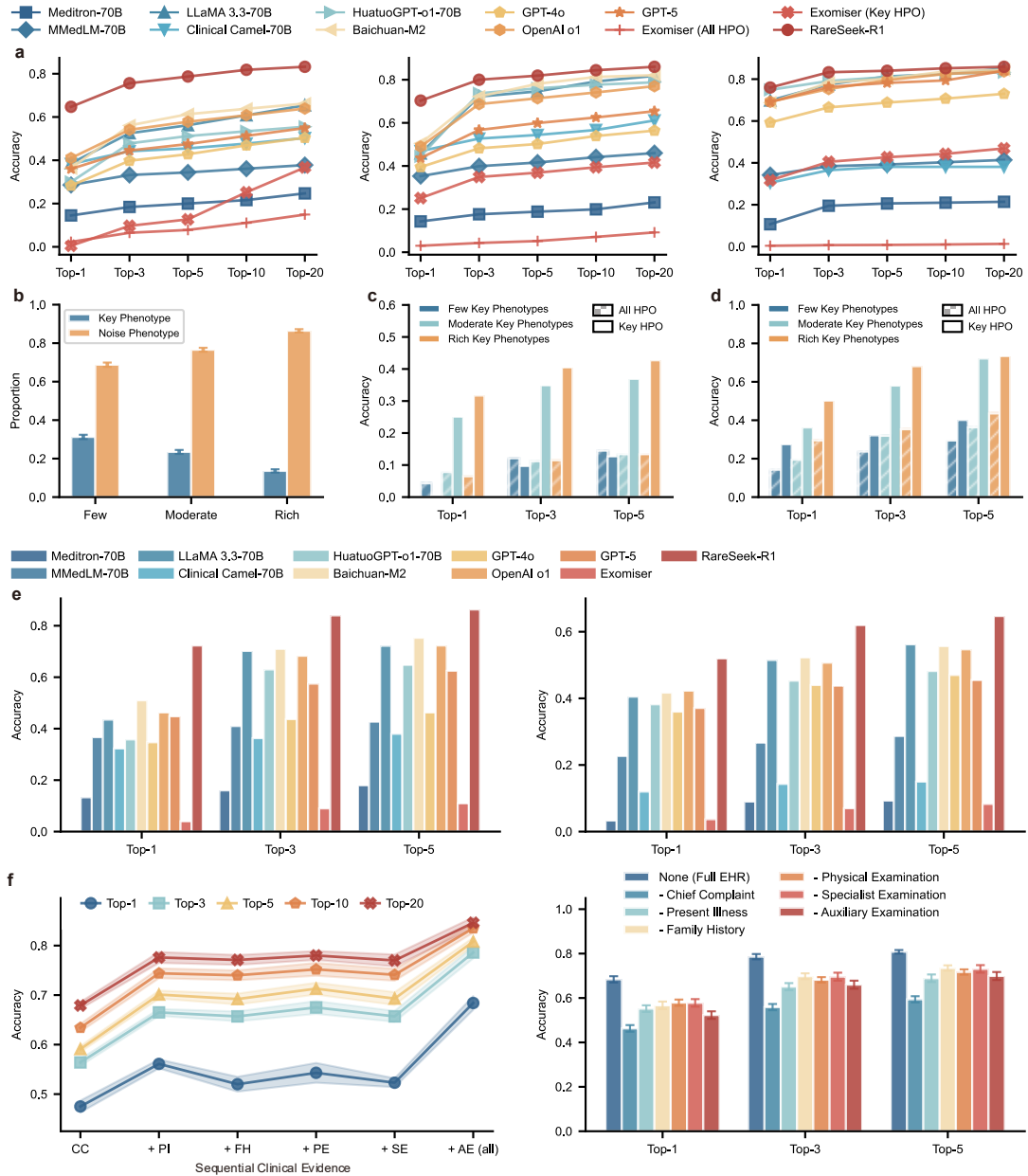

**Fig. 3: Performance analysis under phenotypic noise, complexity, and sequential clinical evidence in rare disease diagnosis.** (a) Diagnostic performance of different methods across the Few (left), Moderate (middle), and Rich (right) key phenotype strata. (b) Proportion of key versus noise phenotypes across Few, Moderate, and Rich key-phenotype groups. (c-d) Comparison of diagnostic performance between Exomiser (left) and RareSeek-R1 (right) under Few, Moderate, and Rich key-phenotype strata using all and key phenotypes. (e) Diagnostic performance comparison for the 300 cases with the lowest Mean IC (left) and the 300 complex cases with the highest composite difficulty scores (right). (f) Effects of sequential addition of information on diagnostic accuracy (left) and contributions of individual clinical evidence

categories evaluated through single-field ablation experiments (right).

**Enhancing diagnostic accuracy through integration of augmented retrieval and genetic information**

To mitigate the limitations of LLMs that rely solely on parametric knowledge and to minimize hallucinations, we integrated a graph-based RAG framework. GraphRAG injects structured variant-gene-phenotype-disease relations, harmonized from widely used, up-to-date resources including ClinVar[8], HGMD[31], HPO[7], OMIM[32], and Orphanet [33], directly into the model's reasoning pipeline (see details in Methods). By supplying precise, contextually relevant knowledge, GraphRAG enables faithful traversal of biomedical links, reduces retrieval noise, and improves diagnostic accuracy in rare disease cases. Beyond HPO-based phenotypes, genetic findings and putative pathogenic variants constitute core clinical evidence in contemporary rare disease workflows. As whole-genome sequencing (WGS), long-read sequencing, and ongoing discovery of disease genes/variants expand the knowledge landscape, timely graph updates further amplify GraphRAG's contribution to LLM reasoning. Accordingly, we evaluated whether integrating GraphRAG into RareSeek-R1 enhances diagnostic performance in cases with complete clinical evidence, encompassing both comprehensive phenotypic profiles and prioritized genetic variant data (see details in Methods).

On the MedEHR-Variant benchmark, which comprises 147 cases pairing full EHRs with prioritized putative pathogenic variants from trio whole-exome sequencing, we evaluated the added value of GraphRAG and the integration of genomic evidence. Using Exomiser as a reference, phenotype-only Top-1 accuracy was $0.129 \pm 0.029$ and increased to $0.408 \pm 0.083$ after incorporating variants, confirming the incremental utility of genetic data. We then quantified GraphRAG's contribution within RareSeek-R1 across input modalities. With EHR-only input, Top-1 accuracy rose from $0.312 \pm 0.060$ to $0.332 \pm 0.085$. With variant-only input, the gain was larger, from $0.477 \pm 0.070$ to $0.557 \pm 0.084$. The greatest improvement occurred when phenotypic and genomic evidence were combined, with Top-1 accuracy increasing from $0.575 \pm 0.057$ to $0.770 \pm 0.055$, underscoring both the incremental benefit of GraphRAG and the synergistic boost from integrating phenotypes with variants (Fig. 4a).

On the Phenopacket-Store benchmark[28], which includes 5,213 GA4GH conformant case level records with harmonized HPO annotations, variant metadata, and reference diagnoses, enabling evaluation of phenotype and genotype aware pipelines. Because raw VCFs were unavailable, Exomiser was assessed under a phenotype only baseline with Top-1 accuracy of 0.270 ± 0.011. RareSeek-R1 without GraphRAG reached 0.210 ± 0.007, whereas GraphRAG augmentation improved performance to 0.414 ± 0.014, indicating that structured retrieval and ranking enhance phenotype driven diagnosis. With variant only input, GraphRAG likewise raised accuracy from 0.479 ± 0.018 to 0.615 ± 0.019. When both phenotypic and genetic evidence were combined, RareSeek-R1 achieved its highest accuracy, increasing from 0.550 ± 0.016 to 0.734 ± 0.010, with Top-10 reaching 0.910 ± 0.007 after GraphRAG (Fig. 4b). Collectively, integrating GraphRAG into RareSeek-R1 delivers clear gains, with the largest lift when full EHRs or standardized phenotypes are paired with prioritized variant evidence. We propose that graph-grounded retrieval injects explicit, current variant-gene-phenotype-disease links into the reasoning process, reducing hallucinations, improving factual calibration, and preserving knowledge currency. The structured context resolves overlapping phenotypes, aligns candidates with underlying genetic mechanisms, and deprioritizes spurious matches typical of phenotype-only inputs. Empirically, GraphRAG increases accuracy across input settings, with the largest gains in combined EHR and variant scenarios where clinical narratives and genomic signals act synergistically.

**Human-AI comparative evaluation and LLM diagnostic capability framework**

To evaluate RareSeek-R1's contribution to real-world clinical decision-making, we conducted a human-AI comparative evaluation focused on rare neurologic and congenital metabolic disorders. These two disease categories were selected because they encompass diverse clinical manifestations and diagnostic modalities, and are supported by well-established multidisciplinary care systems, thereby enabling rigorous and reproducible evaluation of diagnostic performance. Participating physicians were stratified by clinical seniority into three cohorts (including Junior (n = 3), Middle (n = 3), and Senior (n = 3)), and assessed on 12 representative diseases among neurologic and congenital metabolic disorders using 110 de-identified EHRs. All clinicians independently reviewed the full records with diagnostic labels removed and produced ranked differential diagnoses under identical conditions, without external tools or references (see details in Methods). Under these standardized conditions, RareSeek-R1 achieved a Top-1 accuracy of 0.473 (Fig. 4c), outperforming the Junior group and aligning with the Middle group, thereby situating the model within the range of mid-level clinical performance for these task domains. When augmented GraphRAG, the model's Top-1 accuracy rose to 0.582, approaching Senior-level performance (Fig. 4c). These findings indicate that RareSeek-R1 can serve as a reliable decision-support system across clinician experience levels, and that GraphRAG further enhances diagnostic calibration and reliability by curbing hallucinations and aligning inferences with structured biomedical knowledge.

We further quantified the assistive value of RareSeek-R1 with GraphRAG in clinician workflows. Two weeks after the initial unaided round, nine physicians re-evaluated the same EHRs (110 cases), this time informed by the model's GraphRAG-augmented reasoning outputs. Top-1 accuracy increased consistently across experience tiers: Junior rose from 0.376 to 0.545 ($\Delta = 0.169$), Middle from 0.479 to 0.597 ($\Delta = 0.118$), and Senior from 0.576 to 0.670 ($\Delta = 0.094$) (Fig. 4c). Improvements reflected tighter evidence integration and better-calibrated differentials. For example, a case first labeled Angelman syndrome was corrected to Prader-Willi syndrome after the AI highlighted rapid weight gain with abdominal obesity, absence of familial obesity or diabetes, and hypotonia, consistent with the reference. Overall, GraphRAG-enhanced RareSeek-R1 markedly improves clinician performance across seniority, with the largest gains in Junior physicians, while providing traceable rationales that enable targeted diagnostic revision.

Conventional metrics such as accuracy and text scores like BLEU[51] or ROUGE[52] insufficiently reflect the clinical quality of LLM diagnostic reasoning. To provide a clinically grounded assessment, we developed FINDER (Framework for Inference and Diagnosis Evaluation in Rare Diseases), a Likert-based rubric constructed via literature review and expert consultation that organizes LLM performance into eight clinical dimensions and is applied to both the model's reasoning process and diagnostic outputs (see details in Methods). Using the two datasets described above, RareSeek-R1 significantly outperformed OpenAI o1 in Medical Case Comprehension (3.963 vs. 3.584, $P < 0.001$), Medical Guidelines and Consensus (3.888 vs. 3.697, $P < 0.001$), Sensitivity to Key Clinical Features (4.067 vs. 3.729, $P < 0.001$), and Clinical Reasoning (4.003 vs. 3.850, $P < 0.001$), indicating more accurate capture and integration of critical evidence. RareSeek-R1 also scored higher for Differential Diagnosis Relevance (4.002 vs. 3.933, *P* < 0.001) and Diagnostic Acceptability (4.019 vs. 3.773, *P* < 0.001), reflecting better clinical alignment, and for Bias and Fairness (3.923 vs. 3.764, *P* < 0.001) and Potential for Harm (3.954 vs. 3.769, *P* < 0.001), indicating improved safety and reliability (Fig. 4d).

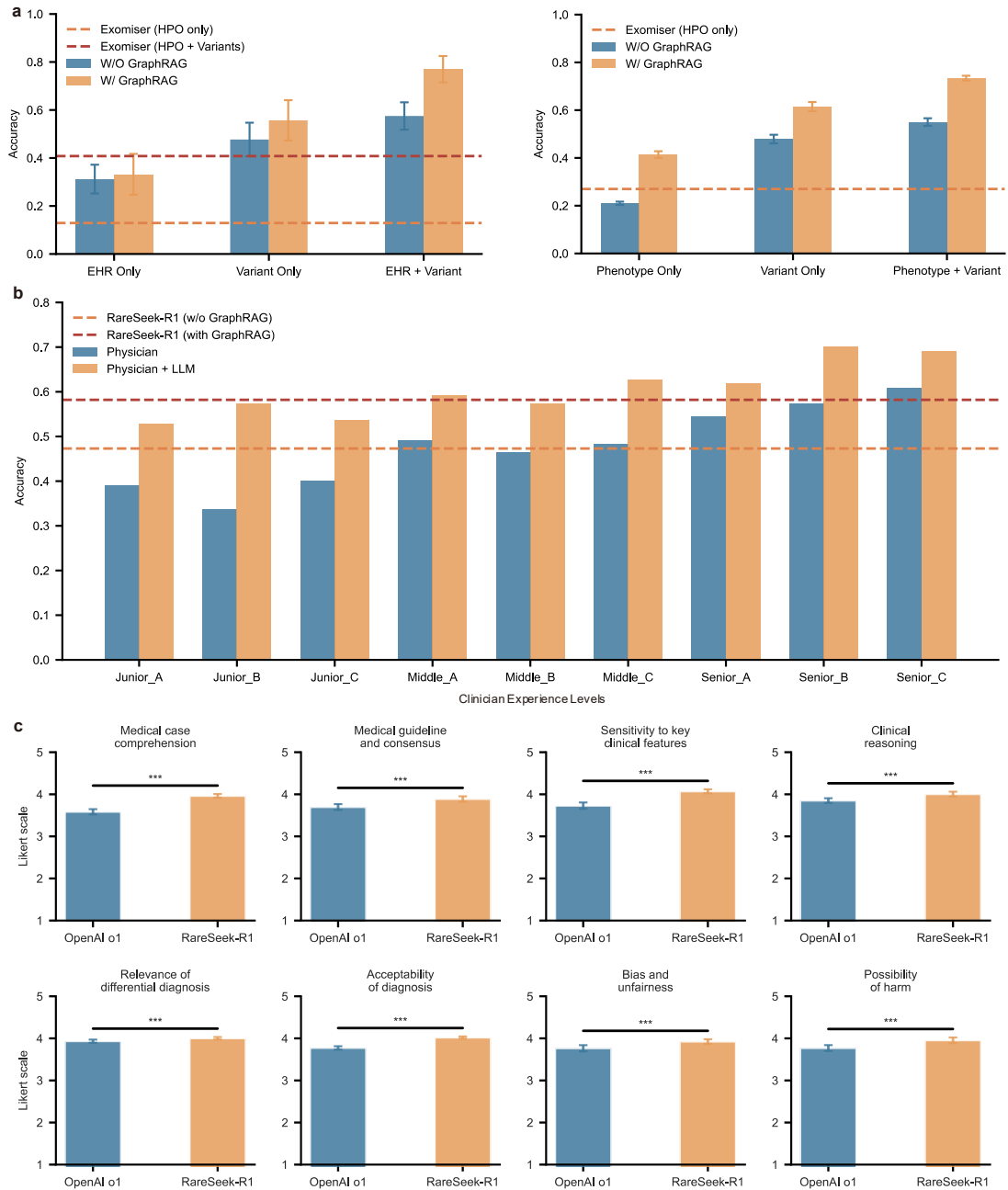

**Fig. 4: Diagnostic performance integrating GraphRAG and genetic information, and comparison of AI-assisted diagnosis with human physicians. (a)** Impact of GraphRAG-based retrieval and genetic information on diagnostic performance. Left: MedEHR-Variant

dataset; right: Phenopacket-Store dataset. **(b)** Diagnosis comparison between AI and clinicians, and effect of AI assistance on clinicians for rare neurologic and congenital metabolic disorders (n = 110). **(c)** Human evaluation of RareSeek-R1 versus OpenAI o1 across eight clinical dimensions. Bar graphs represent mean ± 95% confidence intervals. Statistical significance was determined using two-sided t-tests: ***$P < 0.001$.

**Model interpretability and contribution of non-HPO clinical evidence in RareSeek-R1 model**

To assess interpretability in rare disease diagnosis, we systematically analyzed RareSeek-R1's case-specific CoT reasoning, which provides a stepwise exposition of diagnostic logic and elucidates mechanisms underlying predicted outcomes. We selected four representative disorders from the EHR-Internal dataset, including Wilson disease (ORPHA:905), Prader-Willi syndrome (ORPHA:739), Tuberous sclerosis complex (ORPHA:805), and Langerhans cell histiocytosis (ORPHA:389), and we applied word cloud visualization to highlight high-frequency clinical features emphasized during reasoning. These disorders had sufficient well-documented cases within the EHR-Internal dataset, allowing comprehensive assessment of the model's reasoning interpretability and underlying diagnostic rationale. RareSeek-R1 consistently prioritized clinically salient cues concordant with established literature and disease definitions. For example, low ceruloplasmin and hepatic dysfunction in Wilson disease, and cortical tubers and seizures in Tuberous sclerosis complex, demonstrating precise attention to core diagnostic characteristics. To further probe RareSeek-R1's disease-specific decision-making, we built a global surrogate model using extracted HPO-annotated phenotypes as features and the model's predicted diagnoses as targets. Feature importances derived from regression coefficients reliably recapitulated disease-defining signals. For Wilson disease, the top contributors were elevated hepatic transaminases, jaundice, vomiting, hepatitis, and Kayser-Fleischer ring. For Tuberous sclerosis complex, the leading features were seizure, subependymal nodules, sleep disturbance, status epilepticus, and cardiac rhabdomyoma, each concordant with HPO annotations and canonical disease profiles.

In analyzing RareSeek-R1's reasoning process, we found that HPO-unmappable, non-phenotypic clinical evidence, alongside standardized phenotypes, is pivotal for rare disease inference. For example, Sildenafil therapy, commonly used to treat pulmonary arterial hypertension, cannot be directly mapped to HPO terms, yet it provides valuable diagnostic clues for rare diseases such as Idiopathic/heritable pulmonary arterial hypertension. To quantify this contribution, we examined all correct diagnostic CoT outputs in EHR-Internal, extracting diagnostic features and mapping them to HPO, and unmapped items were annotated as non-HPO evidence (see details in Methods). Overall, the median per-patient proportion of non-HPO features was 23.1% (interquartile range [IQR] 0.118-0.400; Fig. 5a). Among correctly diagnosed conditions with more than 30 cases, non-HPO features averaged over 40% in Langerhans cell histiocytosis, retinoblastoma, and neuroblastoma, with IQRs of 0.400-0.600, 0.375-0.600, and 0.297-0.586, respectively (Fig. 5c). By Orphanet category, averages exceeded 30% in rare gastroenterologic, respiratory, hematologic, ophthalmic, and neoplastic diseases, with IQRs of 0.239-0.441, 0.200-0.533, 0.229-0.520, 0.250-0.500, and 0.300-0.583, respectively (Fig. 5d). These findings underscore that non-HPO evidence is substantial and clinically consequential, and can be effectively leveraged by LLMs to enhance diagnostic reasoning.

We further conducted a systematic categorization of non-HPO features extracted from EHR narratives. Leveraging DeepSeek-R1 for semantic classification (see details in Methods), these signals predominately mapped to seven categories: Imaging findings (26.5%), Clinical interventions or procedures (23.0%), Functional assessments (17.7%), Laboratory test results (9.1%), Genetic or molecular test results (8.2%), Pathology or histology results (7.2%), and Environmental, lifestyle, or exposure factors (5.4%) (Fig. 5b). This distribution indicates that rare disease inference with LLMs hinges not only on standardized phenotypes but also on rich, non-phenotypic clinical evidence that complements HPO-mapped features and frequently carries decisive signal, particularly in oncology, ophthalmology, and multisystem disorders. In complex presentations, integrating both phenotypic and non-phenotypic inputs aligns more closely with clinician reasoning, improves calibration of differential diagnoses, and enhances interpretability through traceable evidence categories. These findings also motivate future AI-enabled rare disease pipelines to incorporate multi-source clinical data and structured categorization of non-HPO evidence, thereby strengthening robustness, clinical validity, and real-world applicability.

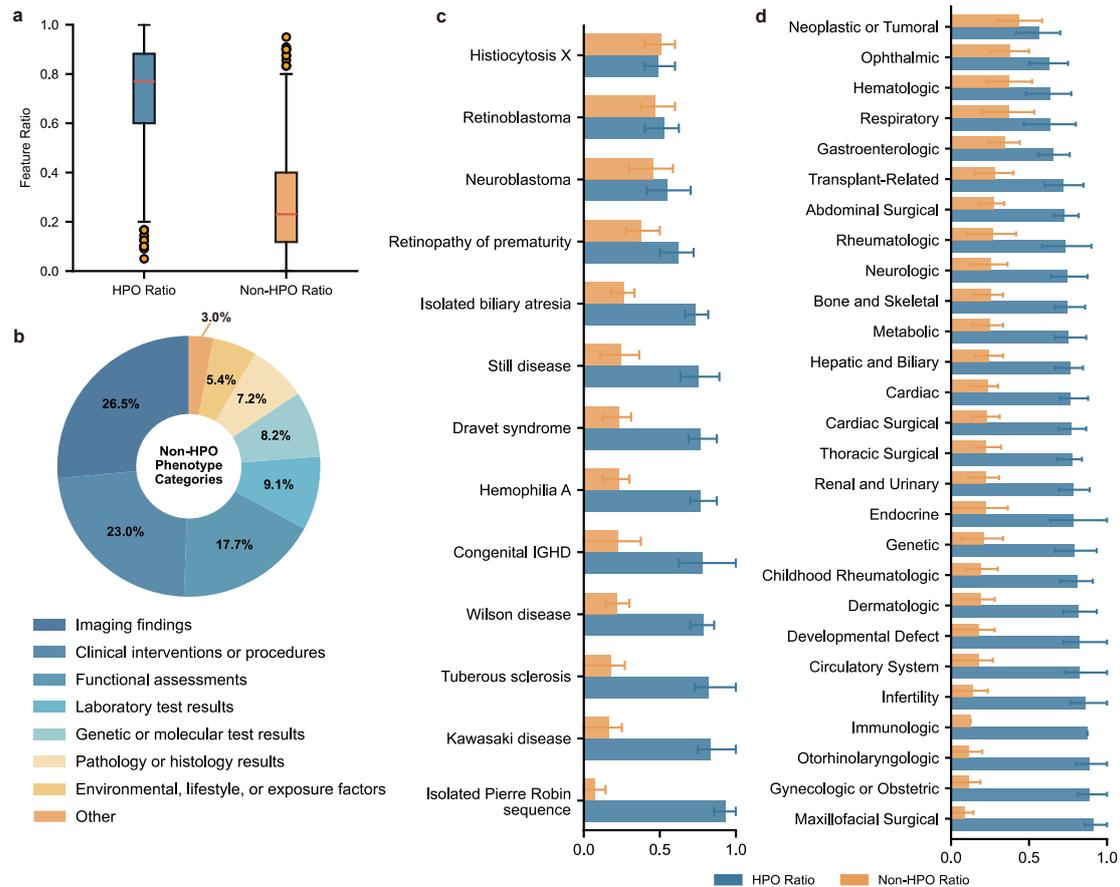

**Fig. 5: Non-HPO clinical evidence of RareSeek-R1 in rare disease inference. (a)** Proportion of phenotype versus non-phenotype features per case (median and interquartile range); **(b)** Distribution of non-phenotype features across categories in the full chain-of-thought reasoning dataset; **(c)** Distribution of non-phenotype features by category in representative disease cases; **(d)** Proportion of phenotype and non-phenotype features across rare disease system categories in representative cases.

## Discussion

We present RareSeek-R1, a domain-specialized LLM for rare disease diagnostic reasoning trained via Progressive Parameter-Efficient Transfer Learning, pairing instruction tuning on a large, clinically grounded RareMed-Corpus deeply integrated from diverse texts and real-world EHRs with high-fidelity fine-tuning on RareMed-CoT to cultivate explicit, stepwise clinical reasoning. Across six internal and external independent benchmarks, RareSeek-R1 delivered consistently strong differential diagnosis performance under heterogeneous case distributions.

Then, human-AI comparisons positioned the model between middle- and senior-level clinicians, and AI-assisted rereads improved accuracy across all tiers, most notably for junior physicians. GraphRAG further reduced hallucinations, improved factual calibration, and aligned inference with latest variant-gene-phenotype-disease relations, with the considerable gains when EHR narratives were combined with prioritized variants. Finally, interpretability analyses confirmed clinical face validity by showing that case-specific CoT traces and a global surrogate model recovered that non-HPO evidence, such as imaging findings, laboratory tests and other clinical procedures, forms a diagnostically consequential component of the model's rationale. Together, RareSeek-R1 reasons directly over full EHR text, leverages graph-grounded retrieval, and measurably augments clinician performance, advancing reliable and scalable AI support for rare disease diagnosis.

Prior work shows LLMs degrade on phenotype-only inputs versus full narratives, indicating that phenotype extraction alone underestimates diagnostic potential[27,53]. Curated benchmarks like RareBench privilege tidy HPO profiles while omitting decisive non-HPO signals, like disease trajectory, treatment response, and laboratory or imaging findings[54,55], thus favoring phenotype-driven tools in artificial settings. In routine care, etiologies are uncertain[56], phenotypes are noisy or overlapping[35,45,57], and multimorbidity is common[58], demanding systems that synthesize heterogeneous evidence rather than operate on phenotypes in isolation. RareSeek-R1 reasons directly over full EHR narratives and preserves non-phenotypic information that phenotype pipelines compress or discard, maintaining superior accuracy under phenotypic noise and complexity where conventional tools degrade. Sensitivity analyses show that even within few key phenotypes, substantial diagnostic signal lies outside HPO-A, explaining consistent gains with narrative inputs. These capabilities both upgrade and disrupt practice. As an upgrade, RareSeek-R1 transforms messy, incomplete notes into actionable differentials, reducing dependence on perfect HPO normalization and mitigating failure modes from ontology gaps, term granularity, and context loss. Operationally, RareSeek-R1 acts as a front-door assistant surfacing high-yield tests/referrals and a back-end auditor reconciling narrative, phenotype, and genomic evidence, shortening the diagnostic odyssey and improving equity. These results support a paradigm shift from phenotype-centric pipelines to narrative-first, graph-grounded clinical reasoning that better mirrors real-world decision.

Although RareSeek-R1 delivers high diagnostic accuracy and interpretability, performance may decline in ultra-rare disorders or in presentations with substantial phenotypic overlap that can be mistaken for common diseases, reflecting limited training data, ambiguous clinical signals, and non-deterministic factors[59,60]. KGs provide structured priors linking diseases, phenotypes, genes, and variants, compensating for long-tail gaps and constraining candidates by genetic and molecular mechanisms to efficiently narrow the search space[55,61]. For example, KCNQ2-related developmental and epileptic encephalopathy (ORPHA:439218) and STXBP1-related encephalopathy (ORPHA:599373) share global developmental delay (HP:0001263), seizures (HP:0001250), and intellectual disability (HP:0001249), making phenotype-only discrimination difficult. Thus, integrating graph-based genetic context removes mechanism-inconsistent candidates and speeds convergence to the correct diagnosis. In practice, this both upgrades existing workflows and enables a more disruptive, narrative-native, graph-constrained approach to end-to-end reasoning. Overall, combining semantic reasoning with KG priors improves accuracy and reliability and underscores the distinctive value of GraphRAG for complex rare disease diagnosis[61].

Interpretability and reliability are prerequisites for clinical use[62-64]. RareSeek-R1 natively produces explicit diagnostic reasoning chains, offering transparent insight into how conclusions are reached. The highlighted cues in its outputs align with established knowledge, for instance, copper accumulation and liver dysfunction supporting Wilson's disease[65], and cryptorchidism with developmental delay supporting Prader-Willi syndrome[66]. Feature relevance from a global surrogate model was concordant with HPO-based disease-phenotype associations, providing convergent validation of mechanism-consistent reasoning. Beyond phenotypes, the chains incorporate non-phenotypic evidence, such as imaging findings (intraocular mass on CT) and genetic data (A3243G mtDNA mutation), that match known disease characteristics. Clinically, this dual validation (reasoning trace plus external priors) both upgrades current workflow transparency and auditability and enables a more disruptive pathway in which narrative-native reasoning is continuously checked against structured knowledge.

RareSeek-R1 adds practical value across care and research while enabling a potential shift in workflows. For clinicians, explicit reasoning chains streamline information capture, integrate multimodal data, and guide interpretation, improving decision quality and auditability. In preclinical triage, it ranks likely diagnoses from narratives and history, and recommends targeted tests or referrals, helping primary care identify complex rare diseases early and coordinate multidisciplinary care. For patients with multisystem or ambiguous presentations, it provides cross-disciplinary support that unifies diverse evidence, and in remote or resource-limited settings it improves early access through personalized testing and referral pathways, potentially reducing delays. For researchers, structured reasoning outputs and a graph-grounded knowledge base aid discovery of disease, phenotype, gene, and variant links, informing mechanistic studies and trial design. Collectively, these roles strengthen clinical management and open a path toward narrative-first and graph-constrained diagnostic workflows that are auditable, updatable, and scalable.

This study has limitations that reflect enduring challenges in the rare disease domain. Over half of rare diseases lack known etiologies, yet training and evaluation labels largely reflect clinical diagnoses rather than etiologic or molecular confirmations, constraining etiologic inference and gene level prediction that are essential for sequencing optimization and discovery of phenotype gene associations[6,67-69]. The current focus on textual and structured data without comprehensive integration of imaging and laboratory signals may limit applicability in complex scenarios where multimodal synthesis is required[55,70-73]. RareSeek-R1 also lacks explicit exclusion mechanisms for common diseases, a choice that may increase false negatives in rare disease detection while still aiding differential diagnosis. In addition, the knowledge landscape evolves rapidly, and existing graphs incompletely capture newly described entities and phenotypic traits, creating gaps in coverage and update cadence. Notwithstanding these constraints, RareSeek-R1 demonstrates robust performance in cases with established etiologic or molecular diagnoses, indicating alignment with clinically meaningful mechanisms. Future work will expand cohorts with etiologic confirmation to strengthen inference at the mechanism level, integrate multimodal evidence across narratives, genetics, imaging, and laboratories to improve performance in complex presentations, thereby advancing RareSeek-R1 from clinical level interpretation toward etiologic level understanding and improving readiness for real world deployment.

## Methods

### Data collection

#### 1. Domain-specific instruction tuning dataset

To enable domain-specific knowledge infusion for RareSeek-R1, we constructed the RareMed-Corpus, a multilingual rare disease dataset comprising four component datasets, RareMedEHR, RareMedText, PubMed-CR, and RareMedSynthetic, which collectively contain approximately 500 million tokens (Fig. 1a).

RareMedEHR comprises 48,852 de-identified electronic health records collected from multiple hospitals across China, including Guangzhou Women and Children's Medical Center (GWCMC) and the Tianjin Healthcare Big Data Super Platform. The dataset encompasses 263 rare diseases, with patients having a mean age of 3.83 years (SD = 5.42), reflecting the predominantly pediatric nature of the cohort. GWCMC, as the largest women and children's hospital in South China and the first in the region with Level-7 EHR capability, contributes a large volume of rare pediatric disease cases, providing extensive longitudinal EHR data for early diagnosis and comprehensive disease characterization. The latter aggregates clinical and public health data from 43 tertiary and 39 secondary hospitals in Tianjin, which, after standardization and de-identification, form an interoperable, research-ready database. The dataset covers the period from January 2016 to April 2024. The study was conducted under a waiver of written informed consent approved by the Institutional Review Board (IRB), and ethical approval was obtained from the IRB and ethics committees of all participating institutions (no.2025[379A01]). All EHR data were rigorously de-identified to remove any patient-identifiable information. Each record includes demographic information, chief complaint, present illness, ancillary tests, family history, physical examination, and specialty assessments. To ensure diagnostic accuracy and consistency, we included only records in which the rare disease diagnosis was explicitly documented in the discharge summary and corroborated by the responsible clinician or by a specialist/multidisciplinary review. For patients with multiple admissions, only the first hospitalization was retained to avoid redundancy and potential bias introduced by longitudinal follow-up. In addition, eligible records were required to contain at least one disease-associated phenotype or equivalent objective diagnostic evidence, including genetic or molecular test results, pathology or histology reports, imaging findings, and other relevant diagnostic evidence. Records were excluded if they lacked essential clinical sections, contained uncertain diagnoses, or provided insufficient information for reliable assessment. Disease names were standardized using the Monarch Initiative v3 API (https://api-v3.monarchinitiative.org/v3/docs#/) to map each diagnosis to the corresponding Orphanet (ORPHAcode) identifier. All mappings were manually reviewed by experienced physicians to resolve inconsistencies. To ensure dataset completeness and cross-source consistency, the data underwent validation, duplicate detection, and random manual review of 5% of the records, achieving an inter-reviewer agreement of $\kappa = 0.91$.

RareMedText was developed as a comprehensive corpus of textual resources on rare diseases, compiled from authoritative medical textbooks, specialty monographs, evidence-based clinical practice guidelines, and disease encyclopedic knowledge sources. The collection comprises 35,722 multilingual entries covering 6,820 distinct rare diseases, systematically encompassing disease definitions, etiologies, clinical manifestations, diagnostic criteria, differential diagnoses, and related clinical knowledge. Source materials were obtained from well-recognized repositories and institutions, including the China Alliance for Rare Diseases (ChARD, https://www.chard.org.cn/), the National Organization for Rare Disorders (NORD, https://rarediseases.org/), Orphanet (https://www.orpha.net/), the Online Mendelian Inheritance in Man (OMIM, https://www.omim.org/), the Chinese Medical Association (https://www.yiigle.com/index), the Rare Disease Wikipedia (https://en.wikipedia.org/wiki), and other authoritative databases across diverse medical disciplines such as pediatrics, genetics, neurology, immunology, and endocrinology. To ensure consistency, all disease names in RareMedText were standardized to ORPHAcode following the same curation and verification process applied to RareMedEHR.

The PubMed-CR dataset comprises 30,101 abstracts and full-text case reports of rare diseases collected from PubMed Central (PMC, https://www.ncbi.nlm.nih.gov/)[74], the largest publicly accessible digital repository of biomedical and life sciences literature. The reports provide detailed descriptions of patients' symptoms, signs, diagnoses, treatments, and follow-up outcomes, with a particular focus on rare, diagnostically challenging, or newly recognized disorders. To identify relevant case reports, we performed systematic searches in PubMed using Orphanet disease names combined with the term "case report" in article titles. All articles and abstracts were sourced from the PubMed Central Open Access Subset (knowledge cutoff: October 2024), with corresponding PMIDs retrieved from the National Library of Medicine database and full texts automatically downloaded. During curation, extraneous content such as references, tables, figures, and supplementary materials was removed, and duplicated or incomplete reports were excluded. All disease names in the extracted reports were subsequently standardized to ORPHAcode, following the same curation and verification process applied to RareMedEHR. This dataset provides a rich source of authentic clinical narratives, enabling the model to learn diverse real-world diagnostic patterns and reasoning processes across complex and low-frequency rare diseases.

The RareMedSynthetic dataset comprises 34,666 phenotype-driven synthetic cases generated from HPO annotations (https://hpo.jax.org/data/annotations) and Orphanet disease-phenotype associations (https://www.orphadata.com/), designed to supplement the representation of ultra-rare or underreported diseases that are insufficiently captured in real-world datasets. For each disease, associated phenotypes were systematically combined to generate plausible case scenarios. To increase the complexity and realism of synthetic cases, additional phenotypes unrelated to the disease were randomly sampled and incorporated as distractor features, simulating noisy or confounding clinical information commonly encountered in practice. Each synthetic case therefore consists of a set of disease-specific phenotypes alongside potential distractor phenotypes, reflecting both characteristic clinical features and the variability of real-world patient presentations. By providing structured and comprehensive synthetic case representations, RareMedSynthetic enables the model to learn richer phenotypic patterns and disease-phenotype associations, particularly for conditions with extremely limited or absent clinical documentation. This synthetic augmentation mitigates data sparsity and enhances the model's capacity to generalize diagnostic reasoning across low-frequency and highly heterogeneous rare disease presentations.

### 2. Chain-of-Thought fine-tuning dataset

To improve the reasoning and generalization capabilities of large language models in rare disease diagnosis, we constructed the RareMed-CoT dataset, a systematically curated resource comprising 17,477 high-quality diagnostic reasoning samples. The dataset construction began with manual annotation of 500 EHRs by clinical experts, which served as a high-fidelity seed dataset[19,75]. For each record, clinicians provided stepwise diagnostic reasoning, documenting key clinical features, laboratory and imaging findings, pathological results, and candidate diseases for differential consideration. Critical decision points and supporting evidence were explicitly recorded to ensure clinical validity, logical rigor, and reproducibility. These seed cases not only captured the complexity of real-world rare disease diagnosis but also served as structured exemplars for subsequent large-scale reasoning generation.

Building on this foundation, 20,000 additional candidate diagnostic reasoning samples were generated using DeepSeek-R1[16], guided by standardized prompts and authoritative rare disease clinical guidelines. The generated dataset spanned 157 distinct rare diseases across multiple organ systems, including endocrine, respiratory, neurological, hematologic, musculoskeletal, and dermatologic disorders, thereby ensuring broad phenotypic coverage and representative clinical scenarios. This strategy allowed the inclusion of both common and ultra-rare disease manifestations, ensuring that the resulting reasoning chains reflect realistic variability in clinical presentations, laboratory patterns, imaging findings, and histopathological characteristics.

To maintain data quality and reliability, we implemented a multi-tiered quality control process. Initially, all generated samples underwent automated pre-screening using ChatGPT-4o[13], which evaluated format integrity, logical coherence, and consistency with established medical knowledge, automatically flagging outputs with evident contradictions or clinical inaccuracies. Samples identified as high-risk or exhibiting residual uncertainty were subsequently manually reviewed by clinical experts, who verified whether each reasoning chain appropriately covered key diagnostic evidence, maintained logical consistency, and avoided critical medical errors. Discrepancies between reviewers were resolved by senior clinical experts, ensuring that only reasoning chains meeting stringent clinical and logical standards were retained. Through this rigorous process, 16,977 high-quality generated reasoning chains were ultimately selected and merged with the original 500 manually annotated seed cases to form the final RareMed-CoT dataset[75].

This comprehensive dataset provides a robust training resource for chain-of-thought (CoT) fine-tuning, enabling large language models to move beyond purely outcome-oriented predictions toward process-explainable diagnostic reasoning. By capturing the stepwise clinical logic applied in real-world rare disease diagnosis, the dataset enhances model interpretability, reliability, and generalization across diverse clinical scenarios, supporting downstream applications in both research and clinical decision support systems.

### 3. Evaluation dataset

To systematically evaluate the rare disease diagnostic performance of LLMs against traditional phenotype-driven approaches in real-world clinical settings, we constructed a comprehensive evaluation dataset comprising 11,429 rare disease test cases from multiple sources and data types.

EHR-Internal comprises 4,306 rare disease electronic health records covering 75 distinct diseases that were not used in model training and serve as an internal test set to evaluate RareSeek-R1's diagnostic performance. Each record contains comprehensive clinical information, including chief complaints, history of present illness, family history, physical examination, specialty assessments, and ancillary tests such as laboratory and imaging reports. EHR-External comprises 283 cases spanning 21 rare diseases, collected from hospitals within the Tianjin Healthcare Big Data Super Platform that are different from those contributing to the training dataset. This dataset provides an independent evaluation to assess the model's robustness and generalization across different hospitals, despite originating from the same overarching platform. Both datasets adopted the same physician-confirmed and documentation-based diagnostic verification criteria as RareMedEHR, ensuring consistency and reliability of diagnostic labeling. RareBench, a publicly available benchmark, integrates five rare disease repositories, including MME[34] (40 cases, 17 diseases), LIRICAL[35] (370 cases, 252 diseases), HMS[36] (88 cases, 46 diseases), RAMEDIS[37] (624 cases, 74 diseases), and PUMCH_ADM[24] (75 cases, 16 diseases), totaling 1,197 phenotypic-only cases, which are used to evaluate the model's generalization ability on standardized benchmarks.

To investigate the contribution of genetic variation and GraphRAG-enhanced reasoning to diagnostic performance, two additional datasets were included. MedEHR-Variant comprises 147 cases covering 109 rare diseases from Guangzhou Women and Children's Medical Center, each with complete EHRs and WES data, representing the only test set with both full clinical and genomic information. Phenopackets, from the GA4GH Phenopacket-store repository[76], includes 5,213 standardized cases covering 378 rare diseases, each containing patient phenotypes, variant annotations, and clinical diagnoses in a uniform structured format, enabling cross-study and cross-model comparisons. Collectively, these datasets provide a comprehensive evaluation framework for RareSeek-R1, spanning internal and external validation, phenotypic-only benchmarks, and genotype-inclusive cohorts, thereby allowing assessment of both diagnostic accuracy and generalization across diverse clinical and genomic scenarios.

**Model overview**

Here we developed RareSeek-R1, a domain-specialized large language model for rare disease diagnosis, using a Progressive Parameter-Efficient Transfer Learning framework[30]. This two-stage paradigm integrates domain-specific instruction tuning to inject rare disease knowledge[77] and chain-of-thought fine-tuning to cultivate explicit clinical reasoning[75]. The approach preserves the base model's robust language understanding and contextual modeling ability while systematically instilling expert medical knowledge and reasoning patterns. Consequently, RareSeek-R1 achieves interpretable and clinically grounded diagnostic inferences across complex rare disease cases.

### 1. Domain-specific instruction tuning for rare diseases

In this study, we selected DeepSeek-R1-Distill-Llama-70B as the base model for instruction tuning. This model uses the Llama-3.3-70B architecture as the student model and is trained via knowledge distillation from the more powerful DeepSeek-R1-671B teacher model. Compared with the original teacher model, DeepSeek-R1-Distill preserves the core reasoning capabilities of large models while substantially reducing computational and storage costs[16]. In medical applications, particularly for multi-turn reasoning and diagnostic tasks, DeepSeek-R1 has demonstrated robust performance and reliability[78,79], making it an ideal base model for fine-tuning in rare disease diagnosis.

For instruction tuning, we utilized RareMed-Corpus, a multilingual rare disease dataset integrating four components-RareMedEHR, RareMedText, PubMed-CR, and RareMedSynthetic-collectively encompassing approximately 500 million tokens derived from structured and unstructured clinical data, standardized disease descriptions, and curated phenotype-disease associations. The fine-tuning process systematically infused domain-specific rare disease knowledge into the model[77], enabling it to interpret complex clinical instructions, reason over heterogeneous data modalities, and accurately recognize specialized medical terminology. This knowledge integration substantially enhances RareSeek-R1's ability to perform diagnostic reasoning across diverse clinical contexts, including ultra-rare or underrepresented conditions with limited real-world documentation.

**2. CoT fine-tuning for enhanced diagnostic reasoning**

To further enhance the model's diagnostic reasoning capabilities in complex clinical cases, we employed a CoT fine-tuning strategy[75,80]. Unlike direct prediction of final diagnoses, CoT fine-tuning requires the model to explicitly generate clinically coherent reasoning chains prior to producing diagnostic outputs. These chains encompass reasoning steps from symptoms to candidate diseases, interpretation of genetic variants, differential diagnosis logic, and explanatory notes on exclusion processes. This approach better emulates the cognitive workflow of clinicians, improving both interpretability and diagnostic accuracy.

The training data were derived from two sources. First, a small set of clinician-annotated diagnostic process records served as high-quality seed data[75]. Second, leveraging the DeepSeek-R1 teacher model in combination with authoritative rare disease diagnostic guidelines, we automatically generated a large-scale set of diagnostic reasoning samples to expand the training corpus. Each training instance consists of three components: the input patient record $x_i$, the diagnostic label $y_i$, and the corresponding clinician-annotated reasoning chain $r_i$, with reference to the relevant disease guideline $g_i$. Based on this structure, the large language model is trained to generate coherent diagnostic reasoning chains prior to final diagnosis prediction:

$$\hat{r}_i = [(x_i, y_i, r_i, g_i)]_{i=1}^{N}$$

subsequently, the data were organized into standardized prompt-completion pairs:

$$S'_i = (p_i, c_i), \quad p_i = (x_i, g_i), \quad c_i = (\hat{r}_i, y_i)$$

where $p_i$ represents the input prompt encompassing patient case information and relevant clinical guidelines, and $c_i$ denotes the corresponding generated reasoning chain and final diagnosis. Based on this formulation, the fine-tuning procedure first integrates each patient case and associated guideline into the input prompt $p_i$, under which the model generates a complete diagnostic reasoning chain $\hat{r}_i$ and the corresponding diagnosis $y_i$. Supervised learning is then applied by aligning the model outputs with the standardized completion $c_i$, enabling the model to progressively learn the comprehensive reasoning pathway from clinical and genetic information to candidate diseases and differential diagnoses, thereby systematically enhancing its clinical reasoning capabilities. To further improve training efficiency and reduce computational overhead, this fine-tuning process was implemented using Low-Rank Adaptation (LoRA)[81].

**Knowledge Graph construction and retrieval-augmented generation**

To enable structured integration of genetic and phenotypic knowledge for rare diseases, we constructed RareMed-RAG, a domain-specific biomedical KG designed to support downstream diagnostic reasoning, knowledge retrieval, and retrieval-augmented generation (RAG) tasks[82-84]. The KG was implemented using Neo4j, providing scalable storage, querying, and traversal capabilities for heterogeneous entities and relationships. It encompasses four core entity types-diseases, phenotypes, genes, and variants-forming the foundational infrastructure for comprehensive cross-entity reasoning.

Data integration leveraged multiple authoritative biomedical databases and standardized ontology resources. ClinVar[8] (20250209 version) and the Human Gene Mutation Database[31] (HGMD Professional 2024.2) provided variant-level knowledge, including pathogenic variants, genomic coordinates, and clinical significance. After deduplication of overlapping loci between the two sources, a total of 635,439 unique variant sites were retained. OMIM contributed 7,534 entries, supplying phenotype-gene associations that link genetic mechanisms to clinical manifestations[32]. Orphanet provided comprehensive disease definitions, hierarchical relationships among disease entities, and curated disease-phenotype associations annotated with frequency attributes describing their occurrence in patient populations (e.g., always present, very frequent, frequent, occasional, rare), encompassing 4,281 annotated disease-phenotype relationships[33]. These resources collectively ensured disease-level standardization and enriched the integration of complementary knowledge sources. The HPO played a central role in the workflow by standardizing phenotypic terms and integrating multiple internal relational resources, including HPO Annotation (HPOA) files such as disease-phenotype mappings, gene-phenotype associations, and gene-disease links[7]. In total, the HPO-based integration captured 9,625 OMIM-coded diseases, 8,217 Orphanet-coded diseases, 16,694 unique phenotypes, and 5,186 genes, enhancing cross-entity connectivity and facilitating comprehensive linkage across genetic and phenotypic data. All source databases are periodically synchronized, and the KG is regularly updated to incorporate the latest biomedical and genomic information.

The construction workflow followed several sequential stages. First, raw source datasets were parsed and transformed into a unified data schema. Second, entity normalization was performed: disease concepts were harmonized to authoritative identifiers and terms (Orphanet and OMIM), phenotypic terms were mapped to HPO entries, and genes were standardized using HGNC nomenclature. Variants were normalized using transcript-level identifiers along with standardized nucleotide (c.) and protein (p.) notations, and were linked to genomic coordinates where available. Third, semantic relationships were instantiated across resources, including disease-phenotype, disease-gene, gene-variant, and phenotype-gene edges, while supplementary relations from HPO and curated external resources were incorporated to enhance connectivity and coverage. Finally, the integrated KG was imported into Neo4j and subjected to automated and manual quality-control procedures, including relationship consistency checks, redundancy removal, and structural integrity validation, ensuring coherence, reliability, and traceable provenance of the graph.

In the knowledge retrieval stage, this study employed Graph Cypher Retriever as the core retrieval mechanism instead of conventional vector-based retrieval. Traditional vector retrieval methods typically rely on semantic similarity measures, which are prone to introducing substantial noise. For entities requiring high precision, such as genes and variants, this often results in irrelevant or even incorrect matches. In contrast, the Cypher-based retrieval approach, leveraging the query language of the graph database, directly operates on the explicit indices and relational structures within the KG, thereby achieving high determinism and precision in entity localization and relationship extraction. This characteristic aligns well with the stringent requirements of rare disease diagnostic reasoning, providing a reliable and verifiable knowledge foundation for subsequent inference and generation tasks.

The retrieval process proceeds as follows: first, the input query objects (e.g., diseases, phenotypes, genes, or variants) are mapped to standardized identifiers consistent with the KG schema. Second, Cypher queries are dynamically constructed according to the entity type to accurately locate target nodes and their associated relationships within Neo4j. For phenotype-based retrieval, which may yield a large number of candidate diseases, the results are further filtered and ranked using IC metrics to prioritize candidates with higher informativeness and stronger discriminative power. Finally, the processed subgraph is returned in a structured format and incorporated as external knowledge input to the language model, supporting downstream reasoning and generation processes.

**Baselines for rare disease diagnosis**

To comprehensively evaluate the performance of RareSeek-R1, two categories of baseline methods were selected:

**1. Phenotype-driven tools**

We incorporated multiple phenotype-based diagnostic tools, including Exomiser[5], Phen2Disease[46] (both double and patient modes), Base_IC[46], PhenoDP[45], and Phrank[47]. Among them, Exomiser is a widely adopted tool in rare disease research, originally designed for the joint prioritization of genes and variants by integrating phenotypic and sequencing data. In this study, we employed its "phenotype_only" mode to enable disease ranking purely based on phenotypic inputs. Phen2Disease is a recently proposed phenotype-driven diagnostic approach that applies IC-weighted HPO features and a bidirectional patient-disease similarity metric for disease prioritization; we evaluated it under its phenotype-only configuration as well. Base_IC performs ranking by identifying overlapping phenotypic terms between a patient and candidate diseases and summing their IC values. PhenoDP, a deep learning-based approach, generates phenotypic summaries, recommends relevant symptoms, and optimizes disease ranking to improve the diagnostic efficiency of Mendelian disorders. Phrank represents a classical gene and disease prioritization algorithm whose phenotype-only mode extends disease-associated phenotypes and applies Bayesian network-based and information-theoretic scoring to perform ranking.

**2. Large language models**

To systematically assess the diagnostic potential of LLMs in rare disease diagnosis, we selected two presentative categories of models. First, medical LLMs include Meditron-70B[22],

MMedLM-70B[21], Clinical Camel-70B[23], HuatuoGPT-o1-70B[43], Baichuan-M1[44], and Baichuan-M2[20]. These models were pretrained or instruction-tuned on large-scale medical corpora and clinical datasets to enhance medical knowledge coverage and clinical reasoning capabilities. Second, general LLMs, including LLaMA 3.3-70B[38], QwQ-32B[42], GPT-4o[13], OpenAI o1[41], and GPT-5[39], were trained on diverse cross-domain corpora to achieve broad language understanding and generative competence. Their inclusion allows for a systematic comparison between domain-specific and general-purpose models in rare disease diagnostic tasks, highlighting their differences and complementarity. Notably, several models, such as QwQ-32B, Baichuan-M2, and OpenAI o1, incorporate reasoning-enhanced mechanisms within their architectural design or training strategies. These mechanisms aim to strengthen complex logical reasoning and multi-step inference, thereby enabling the evaluation of the effectiveness of reasoning enhancement in improving diagnostic performance under complex clinical scenarios.

**Evaluation of LLM-based diagnostic results**

    **1. Standardization of disease names**

In evaluating the diagnostic performance of LLMs, standardizing disease nomenclature is a prerequisite for achieving objective comparison and automated evaluation. Since model-generated disease names are typically expressed in natural language, direct text matching is susceptible to variations in phrasing and semantic ambiguity, potentially introducing evaluation bias. To address this issue, all model outputs were first processed to extract disease names from the unstructured diagnostic text responses. Given that the outputs of LLMs often contain contextual reasoning, justification, or ranked disease lists, we employed ChatGPT-4o to perform structured disease name extraction. A rule-based prompt template was designed to instruct the model to identify and output only the disease entities mentioned, while ignoring descriptive reasoning or unrelated text. This step ensured that the subsequent normalization process operated on a consistent and unambiguous set of disease terms.

Following extraction, all disease names were mapped to standardized ORPHAcodes. Compared with other ontological systems such as OMIM or MONDO, Orphanet provides both unique identifiers for rare diseases and a comprehensive hierarchical classification system, facilitating subsequent system-level statistical analyses. For implementation, the Monarch Initiative v3 API (https://api-v3.monarchinitiative.org/v3/docs#/) was used to map natural language disease names to their corresponding Orphanet entities. Model-predicted names that could not be mapped to any valid Orphanet entry were excluded to maintain the reliability and reproducibility of diagnostic accuracy evaluation.

### 2. Definition of correct diagnosis

During evaluation, model outputs were compared against the ground truth based on standardized ORPHAcodes. A prediction was considered correct if the model-predicted code exactly matched the true diagnosis. Furthermore, recognizing the hierarchical nature of disease ontologies and the variability in clinical naming conventions, we additionally considered predictions corresponding to the parent term (i.e., a higher-level disease category) of the true diagnosis as correct. For example, if the ground-truth diagnosis was Mixed cryoglobulinemia type II (ORPHA:93554) and the model predicted Cryoglobulinemic vasculitis (ORPHA:91138), the prediction was still regarded as correct because the latter represents the parent concept of the former in the Orphanet hierarchy. For each diagnostic task, the model was required to generate the top 20 most probable disease predictions for subsequent ranking-based evaluation.

**Exploration and analysis of phenotype information in rare disease diagnosis**

Within the aforementioned evaluation framework, the EHR-Internal dataset, derived from the Guangzhou Women and Children's Medical Center, represents the largest collection in terms of both disease diversity and number of cases. Given its representativeness and data sufficiency, this dataset was selected as the primary analytical cohort for subsequent experiments and analyses to ensure the robustness and generalizability of the evaluation results.

### 1. Phenotype extraction

In the domain of clinical natural language processing (NLP), most existing phenotype extraction tools are optimized for English-language corpora. To address this limitation, the original Chinese EHR texts were first translated into English using the DeepSeek-R1 model[16]. Subsequently, three widely used automated phenotype extraction tools, including PhenoTagger[49], PhenoBERT[4], and ClinPhen[50], were applied to the translated texts, each executed with their default parameter configurations. Model selection was guided by downstream diagnostic performance within phenotype-driven diagnostic pipelines. Specifically, the accuracy of rare disease prediction using extracted phenotypes served as the principal evaluation metric, allowing for a comparative assessment of the three extraction tools and the identification of the most effective phenotype extraction strategy.

**2. Diagnostic comparison between EHR and phenotype-based inputs**

To systematically evaluate the impact of different input formats on the diagnostic performance of LLMs, two controlled experimental settings were established. (1) In the EHR-text condition, the original unstructured clinical narratives were directly input into the model, simulating real-world diagnostic reasoning without prior data curation. (2) In the phenotype-only condition, the input consisted of phenotypic features automatically extracted from the EHR text and subsequently standardized using HPO terms. This setting enabled the evaluation of diagnostic performance when LLMs operate on structured, ontology-aligned phenotype representations. For each clinical case, diagnostic predictions were generated under both conditions, and the resulting disease rankings and candidate lists were compared. This design enabled a quantitative assessment of how structured phenotypic representations influence the diagnostic accuracy and interpretability of LLMs.

**3. Evaluation of model robustness under phenotypic noise and complexity**

To investigate how the Noise phenotypic affects the diagnostic performance of LLMs, we further stratified cases according to the number of key phenotypes. Here, key phenotypes are defined as phenotypic features extracted from the EHR text that match the known disease-associated phenotypes annotated in the HPOA dataset, representing features with direct diagnostic relevance. For each case, the number of key phenotypes was quantified and categorized into three levels: Few key phenotypes ($0 \leqslant n \leqslant 1$), Moderate key phenotypes ($1 < n \leqslant 4$), and Rich key phenotypes ($n > 4$). Diagnostic predictions were then generated within each stratum using the LLM, and both disease ranking accuracy and stability metrics were recorded. By comparing performance across these strata, we systematically evaluated how the abundance or scarcity of key phenotypic information affects model behavior. This analysis provides insight into the model's robustness under heterogeneous clinical data conditions and its adaptability to varying levels of phenotypic completeness in real-world rare disease diagnosis.

Building on this analysis, we further examined phenotypic complexity, reflecting the intrinsic heterogeneity and specificity of phenotype distributions in rare disease diagnosis. In rare disease research, IC is commonly employed to quantify phenotype specificity: phenotypic concepts that appear infrequently in the corpus are assigned higher IC values, whereas lower IC values correspond to more general terms within the ontology hierarchy. The average IC metric provides a measure of phenotype sparsity and diagnostic complexity within a dataset, serving as a reference for task difficulty when interpreting model performance. In this study, we first computed the average IC for all cases in the EHR-Internal dataset and selected the Top-300 cases with the lowest average IC as samples representing high-complexity phenotypic profiles. Given that some cases contain a large number of extracted phenotypes, including many irrelevant or noisy features, we further defined a composite difficulty score that integrates both the average IC and the number of phenotypes, normalized to form a unified difficulty metric. In this metric, higher scores indicate greater diagnostic difficulty, which is inversely related to the average IC alone. Using this criterion, the Top-300 cases ranked by composite difficulty were designated as high-complexity samples, enabling the evaluation of model performance in scenarios characterized by both noisy phenotypic inputs and low phenotype specificity.

$$\text{norm\_mean\_IC}(s) = \frac{\text{Mean\_IC}(s) - \text{Mean\_IC}_{min}}{\text{Mean\_IC}_{max} - \text{Mean\_IC}_{min}}$$

$$\text{norm\_cnt}(s) = \frac{n - n_{\min}}{n_{\max} - n_{min}}$$

$$\text{composite\_difficulty}(s) = (1 - \text{norm\_mean\_IC}(s)) \times \text{norm\_cnt}(s)$$

Mean_IC(s): the average information content of disease *s*, computed over all phenotypes annotated for that disease. Mean_IC$_{min}$: the minimum average information content across all diseases in the dataset. Mean_IC$_{max}$: the maximum average information content across all diseases in the dataset. norm_mean_IC(s): the normalized average information content of disease s, scaled between mean_IC$_{min}$ and mean_IC$_{max}$. n: the number of phenotypes associated with disease *s*. $n_{min}$: the minimum number of phenotypes observed for any disease in the dataset. n$_{max}$: the maximum number of phenotypes observed for any disease in the dataset. norm_cnt(s): the normalized phenotype counts for disease *s*, scaled between n$_{min}$ and n$_{max}$. composite_difficulty(s): the composite difficulty score for disease *s*, integrating both the normalized average information content and the normalized phenotype count to quantify overall diagnostic complexity.

**Clinical study**

**1. Sequential contribution of clinical evidence categories in rare disease diagnosis**

In the clinical diagnosis of rare diseases, patient information is typically acquired in a sequential and progressive manner, beginning with the chief complaint, followed by the history of present illness, family history, physical examination, specialist examination, and auxiliary investigations. To systematically evaluate how each type of clinical information contributes to the diagnostic reasoning process of LLMs, we designed two complementary experiments: Incremental Information Addition Analysis and Single-Field Ablation Analysis.

In the incremental analysis, the chief complaint was used as the baseline input. Additional information fields, including history of present illness, family history, physical examination, specialist examination, and auxiliary tests, were progressively added in the natural order of clinical information acquisition. After each addition, model inference was performed to measure the marginal improvement in diagnostic accuracy attributable to the newly introduced information. In the ablation analysis, the full EHR was used as the baseline. In each iteration, one specific field was removed while keeping all others intact, and diagnostic performance was re-evaluated. The relative decline in accuracy upon the removal of each field was used to quantify its importance to the overall diagnostic capability of the model.

In summary, these complementary analyses provide a fine-grained understanding of how structured EHR components contribute to rare disease diagnosis, offering empirical evidence to guide the optimization of input structures and information utilization strategies for LLM-based clinical reasoning.

**2. Comparison of diagnostic accuracy between AI and physicians**

To systematically assess the role of the AI system in real-world clinical decision-making, we conducted a human-AI comparative experiment evaluating diagnostic performance across clinicians with different levels of experience and an AI system integrating RareSeek-R1 with GraphRAG in rare neurological and congenital metabolic disorders. Participating clinicians were categorized into three groups according to their clinical seniority: junior (n = 3), intermediate (n = 3), and senior (n = 3). A total of 110 clinical cases covering 12 representative rare diseases were included for evaluation.

All clinicians independently formulated diagnostic conclusions under identical conditions, based solely on standardized EHR data, including the chief complaint, history of present illness, family history, physical examination, specialist examination, and ancillary investigations, without access to external references or tools. The AI model was provided with the same structured EHR inputs to generate ranked candidate diagnoses, enabling direct comparison of diagnostic accuracy and agreement across clinician groups and the AI system.

### 3. Assisted diagnostic accuracy with the LLM in the workflow

To further evaluate the assistive role of the AI model in clinical diagnostic workflows, we conducted an extended assessment based on the preceding human-AI comparative experiment. Two weeks after the initial independent diagnosis session, all three clinician groups (junior, middle, and senior) re-evaluated the same 110 rare disease cases, this time with access to the model's reasoning rationales and diagnostic suggestions. Each clinician independently reviewed and revised their diagnostic conclusions using the AI-generated outputs as supplementary references. The final diagnoses were then compared with those obtained during the initial unaided phase. By analyzing the change in diagnostic accuracy across clinician groups before and after AI assistance, this study systematically quantified the diagnostic gain conferred by the large language model and elucidated its potential utility as an augmentative tool in rare disease clinical decision-making.

### 4. FINDER: A human evaluation framework for rare disease diagnosis with LLMs

To systematically evaluate the diagnostic capability and potential limitations of LLMs for rare diseases in real-world clinical settings, we developed FINDER (**F**ramework for **In**ference and **D**iagnosis **E**valuation in **R**are Diseases). This framework was designed with reference to expert consensus in rare disease medicine, domain-specific literature, and established LLM evaluation methodologies, aiming to comprehensively assess the model's medical consistency and clinical feasibility in generating accurate and reliable diagnostic outputs[19,51,52].

FINDER provides a multidimensional quantitative evaluation of rare disease LLMs across eight key dimensions, covering case understanding, diagnostic reasoning, and safety considerations: (1) Medical Case Comprehension: evaluates the completeness and accuracy of the model's semantic understanding and extraction of key information from clinical narratives; (2) Medical Guideline and Consensus Compliance: examines whether the model's reasoning and diagnostic conclusions align with established medical guidelines and expert consensus; (3) Sensitivity to Key Clinical Features: measures the model's ability to identify both phenotypic and non-phenotypic features essential for rare disease diagnosis; (4) Clinical Reasoning Consistency: assesses whether the model's diagnostic logic is consistent with clinical reasoning pathways and physician thought processes; (5) Relevance of Differential Diagnosis: evaluates the model's capability to differentiate among multiple candidate diseases and identify the most plausible etiologies; (6) Acceptability of Diagnostic: determined by clinical experts, this dimension measures the practical feasibility and reliability of model-generated diagnoses in real-world clinical settings; (7) Bias and Fairness: assesses potential biases related to age, sex, culture, or ethnicity that may affect diagnostic outcomes; and (8) Possibility of Harm: detects the presence of false, misleading, or potentially harmful information that could contribute to misdiagnosis or clinical risk.

In summary, the FINDER framework provides a standardized and systematic foundation for revealing the comprehensive capabilities of rare disease LLMs in medical understanding, clinical reasoning, fairness, and safety.

**Model interpretability and non-HPO clinical evidence in rare disease diagnosis**

To systematically elucidate the reasoning logic and decision rationale of the model in rare disease diagnosis, we conducted a comprehensive interpretability analysis of its generated diagnostic reasoning process. Given the wide heterogeneity of rare diseases and substantial variability in clinical features, we selected four representative disorders with well-documented phenotypic and diagnostic information-Wilson disease, Prader-Willi syndrome, Tuberous sclerosis complex, and Langerhans cell histiocytosis-as analytical cases to ensure representativeness and interpretability. These diseases exhibit complex, multisystem phenotypes and diagnostic logic, making them well suited for both reasoning visualization and global surrogate modeling analyses.

In this analysis, the model not only generated final diagnostic outputs but also automatically produced complete CoT reasoning traces[75]. Based on these reasoning texts, we conducted a feature-level visualization using word clouds to intuitively illustrate the model's salient attention to key clinical features across different disease types. Furthermore, a Global Surrogate Modeling approach was employed to transparently reveal the relationship between input features and model predictions[85]. Specifically, a linear regression model was used as a surrogate to approximate the diagnostic behavior of the LLM, with phenotypic features extracted from electronic health records and the model's predicted outputs as inputs. By visualizing the regression coefficients of the surrogate model, we identified the phenotypic factors most strongly associated with diagnostic predictions, thereby providing a global-level interpretation of the LLM's reasoning logic and decision mechanisms.

In addition, we systematically investigated the underlying mechanisms through which phenotypic and non-phenotypic features contribute to the model's diagnostic reasoning process. Specifically, for cases in which the model produced correct diagnostic predictions, we employed the DeepSeek-R1 model by providing, as input, the diagnostic reasoning chain of each case, the model's predicted diagnosis, and the corresponding Orphanet disease definition and description. Leveraging its intrinsic reasoning capability and the standardized disease definitions, the model extracted clinically relevant information that substantially contributed to the diagnostic decision-making process. Subsequently, all extracted features were semantically mapped to the corresponding Human Phenotype Ontology (HPO) terms using the HPO Ontology Service (https://ontology.jax.org/api/hp/docs#/default/search). Features that could not be successfully mapped were designated as non-phenotypic features. For these non-phenotypic features, we further applied the DeepSeek-R1 model to perform semantic categorization, classifying them into multiple categories, such as Imaging findings, Clinical interventions or procedures, Functional assessments, Genetic or molecular test results, and other relevant categories. Through this systematic pipeline, we quantitatively analyzed the model's dependency patterns on non-phenotypic clinical information within the diagnostic reasoning process, thereby elucidating the latent mechanisms by which large language models incorporate diverse clinical evidence, beyond standardized HPO phenotypes, into rare disease diagnostic decision-making.

**Implementation**

We implemented large LLM training by integrating LoRA[81] and ZeRO-3 with the DeepSpeed framework[86]. Both the Domain-specific instruction tuning and CoT fine-tuning stages employed the same training strategy, ensuring consistent optimization and parameter-efficient adaptation across the two fine-tuning phases. LoRA reduces the number of trainable parameters by freezing the original pre-trained weights and injecting trainable low-rank decomposition matrices into each layer of the Transformer architecture. Parameter-efficient adaptation was achieved through LoRA with a rank of 8 and $\alpha = 32$ applied to all linear layers. The model was fine-tuned for 3 epochs with a per-device batch size of 4 and a learning rate of 1e-4. Attention computations utilized the FlashAttention implementation to optimize efficiency and stability, and training was performed in bfloat16 precision. A warmup ratio of 0.05 was used to gradually adapt the learning rate at the start of training, promoting stable convergence. For inference, we utilized the vLLM library [87], which offers high efficiency in both memory utilization and computational resource management.

**Statistical analysis**

We evaluated the diagnostic performance of RareSeek-R1 using Top-20 accuracy. For each metric, the mean and standard error were calculated. Confidence intervals (CIs) were estimated using a non-parametric bootstrap procedure with 1,000 resamples. In all analyses, a two-sided P value < 0.05 was considered statistically significant. Comparisons between models or input settings were performed using two-sided t-tests to assess whether differences in diagnostic accuracy were statistically significant across datasets.

**Data availability**

Publicly available data were collected from authoritative biomedical sources, including CHARD (https://www.chard.org.cn/), NORD (https://rarediseases.org/), Orphadata (https://www.orphadata.com/), OMIM (https://www.omim.org/), PubMed Central (https://www.ncbi.nlm.nih.gov/), HPO annotations (https://hpo.jax.org/data/annotations) and the Chinese Medical Association database (https://www.yiigle.com/index), as well as publicly accessible encyclopedic resources (https://en.wikipedia.org/wiki). Additional variant-level information was obtained from ClinVar (https://ftp.ncbi.nlm.nih.gov/pub/clinvar/vcf_GRCh38/)

and the Human Gene Mutation Database (https://www.hgmd.cf.ac.uk/ac/index.php). Real-world EHR data were obtained from clinical institutions with institutional review board approval. Due to privacy regulations and institutional policies, the de-identified EHRs cannot be made publicly available. De-identified subsets for academic, non-commercial research may be provided upon formal request to the corresponding author (Mulin Jun Li), following a defined protocol for data request approval.

**Code availability**

The LLMs were developed and deployed in Python (3.10) using PyTorch (2.6.0). The following standard model libraries were used: numpy (1.26.4), scipy (1.15.2), matplotlib (3.10.1), transformers (4.51.3), tokenizers (0.21.1), vLLM (0.8.4) and DeepSpeed (0.16.5).